\def\year{2020}\relax
\documentclass[letterpaper]{article} % DO NOT CHANGE THIS
\usepackage{aaai20}  % DO NOT CHANGE THIS
\usepackage{times}  % DO NOT CHANGE THIS
\usepackage{helvet} % DO NOT CHANGE THIS
\usepackage{courier}  % DO NOT CHANGE THIS
\usepackage[hyphens]{url}  % DO NOT CHANGE THIS
\usepackage{graphicx} % DO NOT CHANGE THIS
\def\UrlFont{\rm}  % DO NOT CHANGE THIS
\usepackage{graphicx}  % DO NOT CHANGE THIS
\title{Modular Neural Ordinary Differential Equations}
\author{Max Zhu, Jacob Moss, Pietro Lio \\
\small University of Cambridge \\
\{mz406, jm2311, pl219\}@cam.ac.uk
}

\usepackage{mathtools}
\usepackage{amssymb}
\usepackage{amsmath}
\usepackage{booktabs}
\usepackage{multirow}
\usepackage{placeins}

\newcommand{\citet}[1]{\citeauthor{#1} \shortcite{#1}}
\newcommand{\citep}{\cite}

\begin{document}

\maketitle

\begin{abstract}
  The laws of physics have been written in the language of differential equations for centuries. Neural Ordinary Differential Equations (NODEs) are a new machine learning architecture which allows these differential equations to be learned from a dataset. These have been applied to classical dynamics simulations in the form of Lagrangian Neural Networks (LNNs) and Second Order Neural Differential Equations (SONODEs). However, they either cannot represent the most general equations of motion or lack interpretability. In this paper, we propose Modular Neural ODEs, where each force component is learned with separate modules. We show how physical priors can be easily incorporated into these models. Through a number of experiments, we demonstrate these result in better performance, are more interpretable, and add flexibility due to their modularity. 
\end{abstract}

\section{Introduction \& Background}
With the increasing popularity of machine learning, many
techniques have been developed to solve physics-based problems. Recently, Neural Ordinary
Differential Equations (NODEs) \cite{NODE} have been proposed. NODEs are ordinary differential equations
(ODEs), with the differential equations represented by trainable neural networks (NNs):
\begin{equation} \frac{d\textbf{X}}{dt}=\textbf{f}(\textbf{X}, t, \textbf{w}) \label{eq: ode}
\end{equation}
where $\textbf{f}$ is a NN function taking in coordinate vector, $\textbf{X}$, time, $t$, and a vector of learnable weights, $w$, that define the NN. This equation can be solved using numerical integration yielding a trajectory. The loss is calculated and the weights of the network are updated by backpropogating \cite{Backprop} through the ODE solver. Many current models represent $\textbf{f}(\textbf{X}, t, \textbf{w})$ as a single black box NN, which lacks interpretability. 
In this work, we propose Modular NODEs, where $\textbf{f}(\textbf{X}, t, \textbf{w})$ in eq. \ref{eq: ode} is composed of separable modules instead of a single NN. Our aim is to make the learned NODEs more interpretable while increasing training accuracy. This method also allows us to incorporate priors to enforce conserved quantities and respect symmetries. The notation we use is described in Appendix \ref{appendix: Notation}. 

\begin{figure}[t]
  \centering
  \includegraphics[width=\columnwidth]{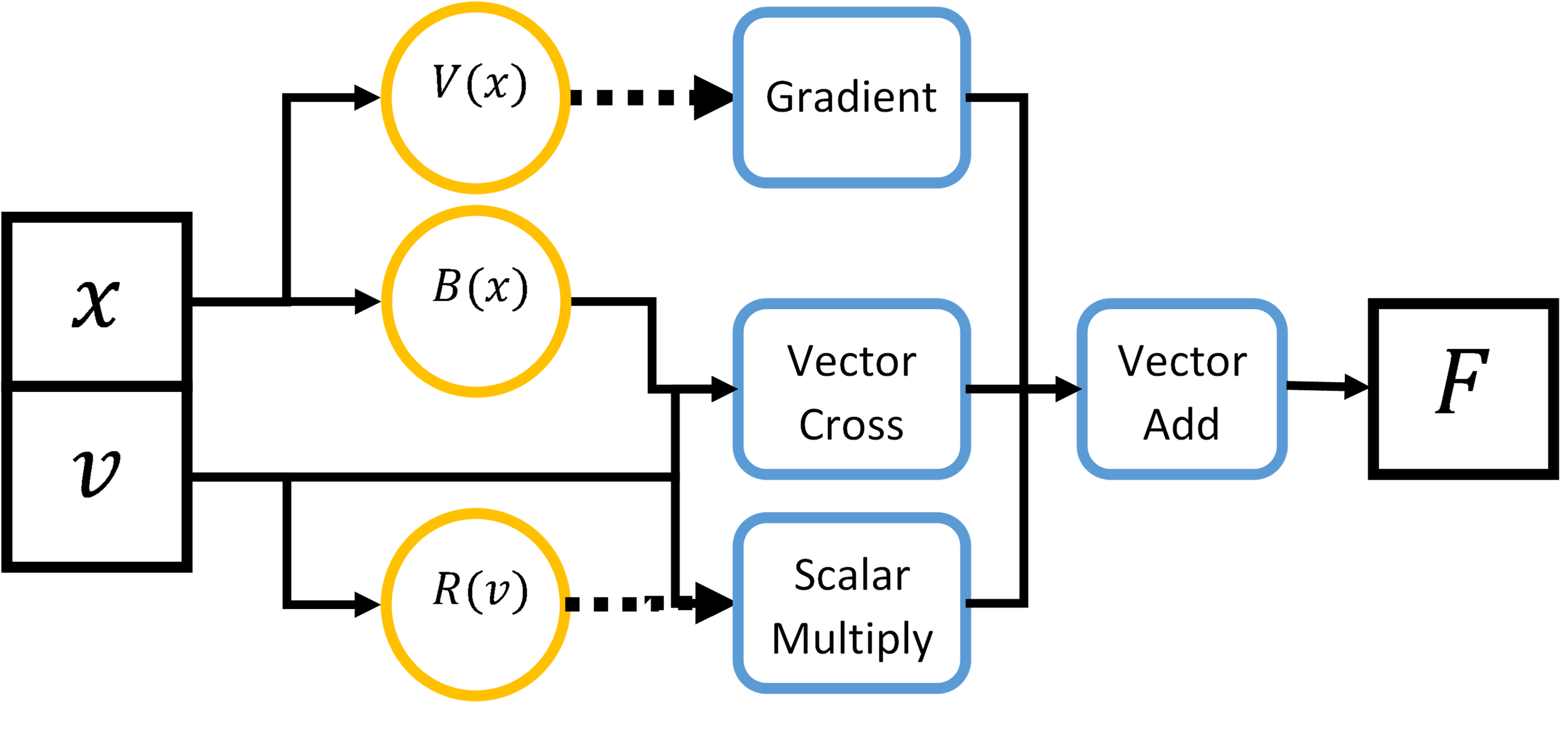}
  \caption{ Structure of a Magnetic NN. Solid lines represent vectors, dotted lines represent scalars, orange circles are trainable neural networks, blue rectangles are vector operations and black squares are input/output vectors. Position \textbf{x} and velocity \textbf{v} are given to V(\textbf{x}),
\textbf{B}(\textbf{x}), and R(\textbf{v}), learnable neural networks that represent potential, magnetic field and drag, respectively. Predetermined
vector operations are applied to the outputs of the neural networks to give the force.}
  \label{fig:Magnetic NN}

\end{figure}
\subsection{Related Work}
\textbf{Second Order Neural Ordinary Differential Equations: }
Work has already been done to improve the performance of NODEs. In Augmented Neural ODEs \cite{ANODE}, additional hidden dimensions allow the NODE to be solved in higher dimensional spaces, allowing more complex functions to be represented. Second Order Neural Ordinary Differential Equations (SONODEs) \cite{SONODE} are a specific case of Augmented Neural ODEs, where the additional dimensions represent velocity coordinates allowing for acceleration to be learned. Newton's Second Law can be expressed as two coupled first order ODEs,
\begin{equation}
\frac{d \textbf{v}}{dt}=\frac{1}{m} \textbf{F} \left(t, \textbf{x}, \textbf{v}\right),\  \frac{d\textbf{x}}{dt}= \textbf{v} \label{eq: coupled ode}
\end{equation}
If we let $\textbf{X}=\left(\textbf{x}, \textbf{v}\right)$ in eq. \ref{eq: ode}, Newtons Law can now be directly learned.

\textbf{Hamiltonian and Lagrangian Networks: }
Hamiltonian Neural Networks \cite{HNN} learn the Hamiltonian of a system, from which forces and equations of motion can be inferred. However, these require specific canonical coordinates to work. Learning the Lagrangian of systems has been investigated in Deep Lagrangian Networks \cite{DeLaN} and Lagrangian Neural Networks (LNNs) \cite{LNN}. Unlike Hamiltonian systems, Lagrangian systems do not need canonical coordinates. It has also been shown that in Lagrangian and Hamiltonian NNs, explicit prior constraints can be included and training in Cartesian coordinates improves accuracy \cite{ConstrainedLNN}. One problem with Hamiltonian and Lagrangian systems is they cannot easily learn dissipative equations of motion.

\textbf{Learning symbolic equations: }
Machine learning has been applied to learning physics in other ways such as AI Feynman 2.0 \cite{AIFeynman}, where symbolic regression is shown to be able to learn symbolic equations from simulated data.  Graph neural networks have been used to learn the dynamics of many body problems and symbolic regression is used to obtain the equations \cite{GNNsymbolic}. The symbolic equations they provide are highly interpretable, but they require large amounts of data and long training times.

\section{Modular Neural Networks} \label{Building Modular NN}
In our model, we propose using a SONODE built from modules made of neural networks (Modular NODEs). Each module learns one or more components of the force, which is decided by how the inputs and outputs of each module are shaped. Several modules are summed together to give a full model which is trained as a NODE. The modules used are predetermined by the types of forces to model. The main advantage is models are more interpretable since each module represents one component of force. This also allows for physical priors to be easily enforced, increasing accuracy and reducing training times. Unlike LNNs and HNNs, they can easily learn dissipative forces (Appendix \ref{appendix: Lagrangian}) and don't require canonical coordinates like HNNs. Using a NODE allows for training directly from positional data.

\begin{table}
  \caption{Summary of types of modules considered here. Functions are learnable neural networks that take in position, $\textbf{x}$ velcity, $\textbf{v}$, or both. The potential model for all Modular NODEs are the same except for the periodic NN, where $\textbf{V} = V(\textbf{x) mod \textbf{a}}$. $V(\textbf{x})$, $D(\textbf{v}), \textbf{B}(\textbf{x})$ and $\textbf{A}(x)$ are potential, drag, magnetic field and vector potential respectively. $\textbf{G}(\textbf{x}, \textbf{v})$ is a more general function. SONODEs and LNNs are also shown.}
  \label{tab: Model Summary}
  \centering
  \begin{tabular}{llll}
    \toprule
    \cmidrule(r){1-2}
    Type                 & Drag                      & Magnetic field    \\
    \midrule
    Magnetic       & $- \textbf{v}D(\textbf{v})$ & $\textbf{v} \times \textbf{B}(\textbf{x})$     \\
    Div Magnetic   & $- \textbf{v}D(\textbf{v})$ & $\textbf{v} \times \textbf{B}(\textbf{x})$, $\nabla \bullet \textbf{B} = 0$    \\
    Vector Magnetic     & $- \textbf{v}D(\textbf{v})$ & $\textbf{v} \times \left(\nabla \times \textbf{A}(\textbf{x} )\right) $ \\
    Basic       & \multicolumn{2}{c}{$\textbf{F} = \textbf{G}(\textbf{x, v})$} \\
    Periodic    & $- \textbf{v}D(\textbf{v})$ & $\textbf{v} \times \textbf{B}(\textbf{x})$     \\    
    SONODE / LNN & \multicolumn{2}{c}{$\textbf{G}(\textbf{x, v})$} \\
    Magnetic LNN & \multicolumn{2}{c}{$L = \frac{1}{2}m\textbf{v}^2 + V(\textbf{x}) - \textbf{v} \bullet \textbf{A}(\textbf{x})$} \\
    \bottomrule
  \end{tabular}
\end{table}

Modular NODEs can be used to describe many different types of forces and many possible configurations of modules. Here, we give demonstrations of the properties and advantages of Modular NODEs using one class of forces, the motion of a charged particle in a potential with a magnetic field and non-isotropic drag in 3D space. Since this is only a demonstration of what Modular NODEs can be used for, there are many other types of forces that may work with Modular NODEs (e.g. more complex drag) but will not be considered here. In Appendix \ref{appendix: Constraints}, we describe constraints to what sorts of modules are usable.

The equation to model is 
\begin{equation} \frac{d^2 \textbf{x}}{dt^2}= -\nabla V(\textbf{x}) + \textbf{B(x)} \times \textbf{v} - f(\textbf{v})\textbf{v} \label{eq: Forces}
\end{equation}
where $V(\textbf{x})$,  \textbf{B(x)} and f(\textbf{v}) are potential, magnetic field and drag respectively. Potential  is always directly learned using a NN module. We experiment with different representations of the potential and magnetic field. In the next sections, we describe the Modular NODEs we create. Names of modules are in \textbf{bold} and a summary is given in Table \ref{tab: Model Summary}.

\subsection{Representations of magnetic fields}

The simplest module we build is the \textbf{Basic NN}, which takes the form $-\nabla V(\textbf{x}) + G(\textbf{x}, \textbf{v})$. However, this is uninterpretable, as we show in section 2.3, and no physical prior is applied. 
A more complex model is the \textbf{Magnetic NN}. This is simply eq. \ref{eq: Forces}, where each of the functions are represented by a NN module taking in the required input vector. The structure is shown in Figure \ref{fig:Magnetic NN}.

To improve the Magnetic NN, we can use the prior that physical magnetic fields always satisfy $\nabla\bullet{\textbf{B}}=\mathbf{0}$ ( Maxwell's Equations \cite{Jacksons}). We incorporate this prior two ways. Firstly, we can add on a loss term

\begin{equation}L_{div}=\left|\nabla\bullet{\textbf{B(x)}}\right|^2,  \label{eq: div loss}
\end{equation}
these will be called \textbf{Divergence Magnetic NN}. $L_{div}=0$ is true everywhere, but this is practically impossible to enforce in a continuous space. Instead, we evaluate the loss at a random point in space at each step. 

Secondly, we can learn the magnetic vector potential $\textbf{A(x)}$, where $\textbf{B(x)}=\nabla\times{\textbf{A(x)}}$. The identity $\nabla\bullet\nabla\times{\textbf{A}}(\textbf{x})=\nabla\bullet{\textbf{B}}=0$ shows $L_{div}=0$ always. NNs which implement this will be called \textbf{Vector Magnetic NN}.

Vector Magnetic NNs are slower to train than Divergence Magnetic NNs, since the derivative $\textbf{B(x)}=\nabla\times{\textbf{A(x)}}$ needs to be backpropagated through the NODE, while the divergence loss can be applied outside the NODE. However, an advantage is that there is no need to introduce another hyperparameter (divergence loss learning rate).

\subsection{Periodic Potentials}
Many systems have potentials that are periodic in space. To enforce this, modular division is used on position coordinates before being given to the potential, $V\left(x\right)\rightarrow V(x\ mod\ a)$ for periodicity a. Any periodic 3D lattice can be simulated this way, but we will only consider simple cubic lattices here for simplicity. We will call these \textbf{Periodic NN}.   

\section{Experiments} \label{section:Experiments}
\subsection{Training Procedure}
 Details about the training procedure, test setups and models are given in Appendix \ref{appendix: Setup}. 

\subsection{Comparing models with magnetic fields}

\begin{figure}
  \centering
  \includegraphics[width=\columnwidth]{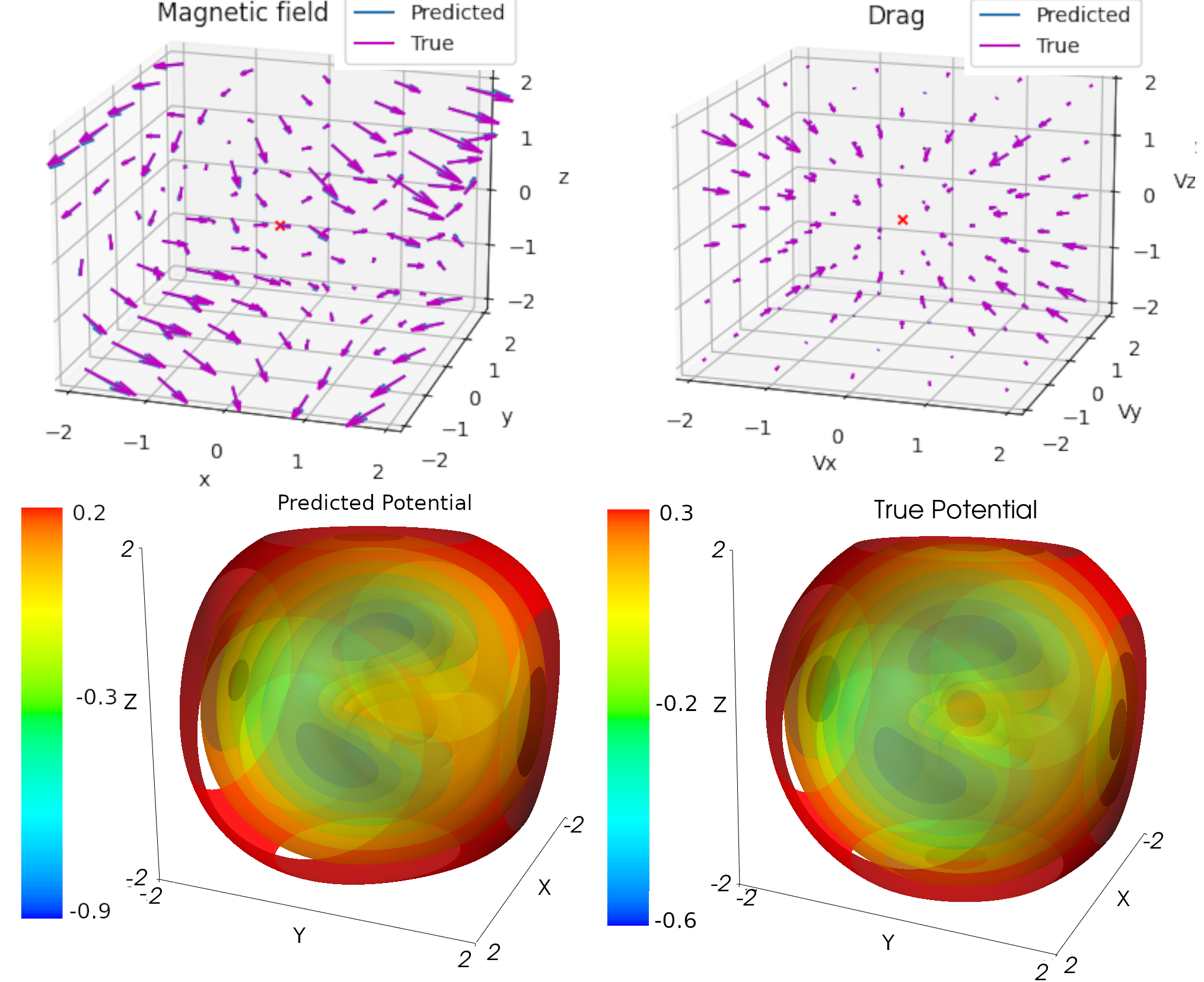}
  \caption{Plots of the learned and predicted magnetic field, drag force and potential from a Magnetic NN. The learned magnetic and drag fields (blue) are nearly identical to the true field (purple).}
  \label{fig:Field Plots}
\end{figure}

Magnetic, Div. Magnetic, Vector Magnetic NN and Basic NN along with SONODEs and LNNs were compared using the \textit{Standard} test setup. In Experiment 1a, the models were trained on a particle moving in a potential with space dependent magnetic fields and velocity dependent drag forces.  
Experiment 1b is the same except the drag force was removed to allow for comparison with the LNN.  We also compare with a standard LNN, the most general unconstrained LNN with no physical priors, and a Magnetic LNN, equivalent to a Magnetic NN without drag written in Lagrangian form (Appendix \ref{appendix: Lagrangian}). To test energy conservation, we also plot the predicted energy over time for a Modular LNN, SONODE, Magnetic NN and a Magnetic NN with drag disabled, using the same setup as Experiment 1b. 

\begin{figure}
  \centering
  \includegraphics[width=\columnwidth]{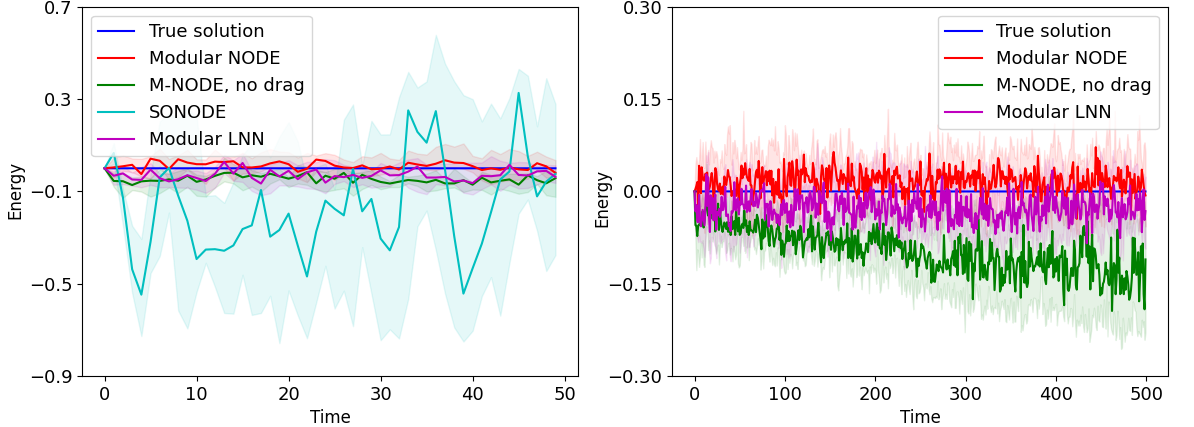}
  \caption{Energy of trajectories measured relative to true potential. \textbf{Left}: shows energy a short time.  \textbf{right}: shows energy for longer times, without the SONODE which diverged in energy. Dark lines are median energy and shaded ares are quartile from 12 models. SONODEs fail to conserve energy, while Modular LNN and Modular NODE without drag conserve energy. The Modular NODE with drag gradually looses energy over time.}
  \label{Energy}

\end{figure}

Results are shown in Table \ref{tab: model comparison}. The modular NODEs and LNNs (without drag) performed the best, while the SONODEs and Basic NN performed the worst. LNNs and Vector NN trained significantly slower. In Figure \ref{fig:Field Plots}, the learned fields from a Magnetic NN are plotted, showing that the modules learn the correct field. 
The predicted energy over short and long periods are plotted in Figure \ref{Energy}. The SONODE and Modular NN do not conserve energy, while Modular LNN and Modular NN without drag conserve energy. Removing the drag in Modular NNs yielded no improvement in the shorter tests we described in Table \ref{tab: model comparison}. 

\begin{table*}
  \caption{ Test mean squared errors after training on a particle in a potential and magnetic field. \textbf{Left:} without drag, and \textbf{right:} with drag. The Magnetic, Divergence and Vector NN, and Magnetic LNN when without drag, performed the best. A SONODE x2 was trained for twice as many steps to match the training time of other methods, but still struggled. }
  
  \label{tab: model comparison}
  \centering
  \begin{tabular}{l|lllp{12mm}|lllp{12mm}}
    \toprule
    &\multicolumn{4}{c}{\textbf{without drag}}& \multicolumn{4}{c}{\textbf{with drag}}\\
    \midrule
    Model & Median & Quartile & Mean  & Train Time (s) & Median   & Quartile     & Mean       & Train Time (s)\\
    \midrule
    Magnetic NN    & 1.27          & 0.73        & 2.54 $\pm$ 0.06  & 363 & 0.46          & 0.24        & 0.84 $\pm$ 0.23  & 391\\
    Magnetic LNN   & 1.46          & 1.18        & \textbf{1.57 $\pm$ 0.23}  & 1332&-&-&-&-\\
    Vector NN      & 1.35          & \textbf{0.80}        & 1.84 $\pm$ 0.34  & 996 & 0.50          & \textbf{0.24}        & 0.68 $\pm$ 0.10  & 973\\
    Divergence NN  & \textbf{1.21}          & 0.80        & 1.85 $\pm$ 0.44  & 398 & \textbf{0.38}          & 0.29        & \textbf{0.50 $\pm$ 0.16}  & 411\\
    Basic NN       & 41.9          & 35.4        & 49.0 $\pm$ 9.14  & 301 & 38.9          & 28.8        & 41.4 $\pm$ 3.94  & 314\\
    LNN            & 46.6          & 34.0        & 49.9 $\pm$ 6.10  & 1556 &-&-&-&-\\
    SONODE         & 95.2          & 45.8        & \textbf{122} $\pm$ 32.5  & 140  & 94.7          & 76.2        & 178 $\pm$ 60.4  & \textbf{165}\\
    SONODE X2      & 89.9          & 47.4        & 176 $\pm$ 51.9  & 342  & 54.6          & 41.7        & 188 $\pm$ 88.7  & 362\\
    \bottomrule
  \end{tabular}
  
  \label{Magnetic drag}
\end{table*}
% \FloatBarrier
Magnetic, Vector and Divergence NNs are compared using the \textit{Magnetic} Testing Setup, with a more complex magnetic field and simpler potential to differentiate between magnetic field representations, in Experiment 2. The three models are tested. Results are shown in Table \ref{Magnetic representations}. We see that the Vector and Divergence NN outperformed the Magnetic NODE, demonstrating the incorporation of the prior $\nabla \bullet{\textbf{B}} = 0$ improved accuracy.

\begin{table}[t]
  \caption{Test mean squared errors when comparing models trained on a simple potential and drag and complex magnetic field. The Divergence NN and Vector NN with magnetic priors build in performed much better. Results are from 16 models each.}
  
  \label{Magnetic representations}
  \centering
  \begin{tabular}{llll}
    \toprule
    \cmidrule(r){1-2}
    Model          & Median         & Quartile     & Mean \\
    \midrule
    Magnetic NN    & 0.125          & 0.065        & 0.163 $\pm$ 0.028 \\
    Div. Mag. NN   & \textbf{0.049}          & 0.\textbf{029}        & \textbf{0.063 $\pm$ 0.009} \\
    Vector Mag. NN & 0.058          & 0.049        & 0.066 $\pm$ 0.015 \\
    \bottomrule
  \end{tabular}
\end{table}

\subsection{Periodic Potentials}
\begin{figure}
  \centering
  \includegraphics[width=\columnwidth]{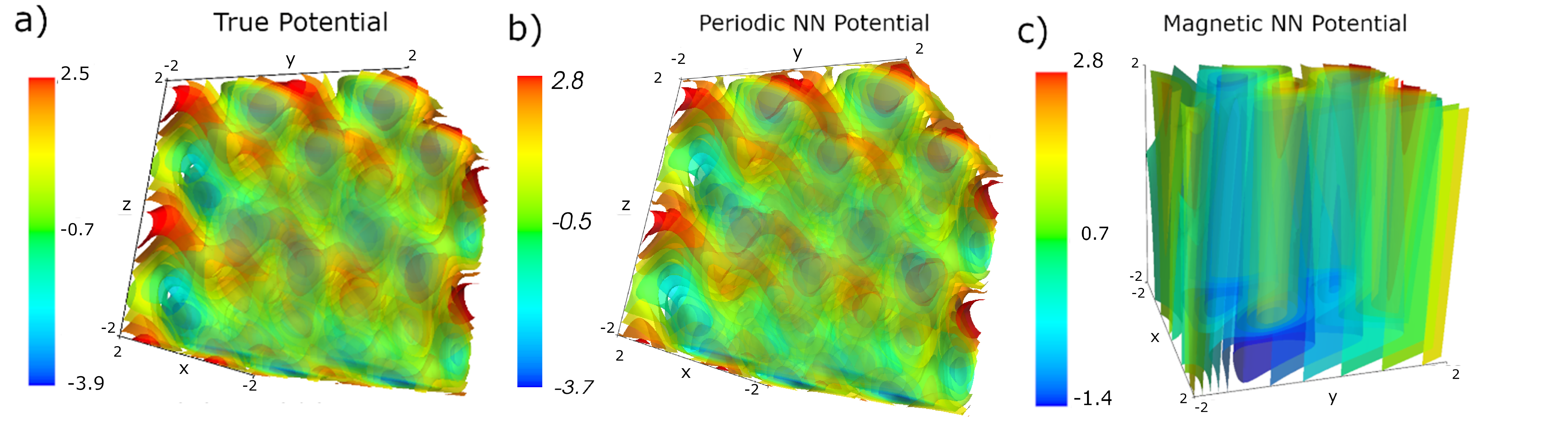}
    
  \caption{Plots of trajectories of models with correct magnetic field, potential field and both. The model with both correct was built from modules from the correct magnetic field and potential, and is the closest to the true trajectory.}
  \label{Periodic}
\end{figure}  
In Experiment 3, Periodic NNs and Magnetic were trained on a particle moving in a periodic potential with a non-periodic magnetic field and velocity dependent drag force. The period of the potential was manually given to the Periodic NN. Results are shown in Table \ref{fig:Periodic results}. The Periodic NN drastically outperformed the Magnetic NN. The reason for the large performance difference can be seen in Figure \ref{Periodic}. The Periodic NN learns a potential that is like the real potential, while Magnetic NN completely fails. 

\begin{table}
  \caption{Test mean squared error of learning the motion of a particle in a periodic potential. The model with periodicity built in performed much better, with little increase in training resources}
  \label{fig:Periodic results}
  \centering
  \begin{tabular}{llllp{12mm}}
    \toprule
    \cmidrule(r){1-2}
    Model          & Median         & Quartile     & Mean               & Train Time (s)\\
    \midrule
    Periodic NN    & \textbf{0.71}          & \textbf{0.37}        & \textbf{1.00 $\pm$ 0.21}    & 407\\
    Magnetic NN    & 70.0          & 38.0        & 142 $\pm$ 37   & \textbf{349}\\
    \bottomrule
  \end{tabular}
\end{table}

\subsection{Combining Learned Modules}
\begin{figure}[t]
  \centering
  \includegraphics[width=0.9\columnwidth]{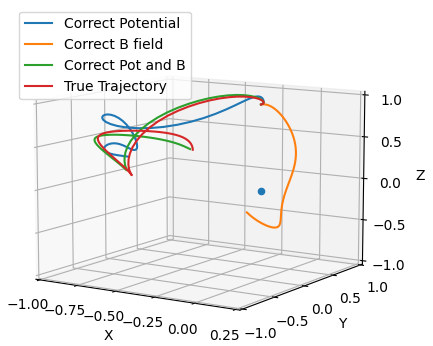}
    
  \caption{Plots of trajectories of models with correct magnetic field, potential field and both. The model with both correct was built from modules from the correct magnetic field and potential, and is the closest to the true trajectory.}
  \label{Combined}

\end{figure}

In Experiment 4, we demonstrate that the learned modules can be used individually. Two models are trained: $\mathcal{M}_A$ uses the \textit{Magnetic} Testing setup, with a complex magnetic field and simple harmonic potential; and $\mathcal{M}_B$ uses the \textit{Standard} Test Setup, with a simpler magnetic field and more complex potential. The drag force is unchanged, although there is no reason this cannot be swapped for another module. The magnetic field and drag forces from $\mathcal{M}_A$ are combined with the complex potential in $\mathcal{M}_B$, yielding $\mathcal{M}_C$, which should be able to model a particle moving in the complex potential and complex magnetic field. 
The results are shown in Figure \ref{Combined}. The modules with only correct magnetic fields and potentials individually performed poorly, the combined  model with correct learned field and potential, is much more accurate. This shows that learned modules are individually interpretable and usable.

\section{Conclusion}
We have shown that Modular NODEs can learn interpretably learn equations of motion. If the force is inherently modular, a higher accuracy can be achieved compared to other NODE based techniques such as SONODEs and LNNs. Priors such as lattice periodicity or Maxwell's equations, can be incorporated into the model in various was to further increase accuracy. Like SONODEs, Modular NODEs can represent both dissipative and conservative forces, but unlike SONODEs, allows for full control of which of these to represent.
However, in order for Modular NODEs to work, the form of the force needs to be known before training. Modular NODEs may not work well for some types of forces where the modules might overlap, leading to degeneracies.  

Further work may include more complex coordinate transforms, such as Fourier Transforms. Furthermore, the correct coordinate transform may help with removing coordinate singularities. 
Another improvement is instead of manually defining the structure of the NODE beforehand, the NODE can be made to learn the correct internal representation of forces. This could be achieved by incorporating a loss term forcing the NN to take of a certain form or to take specific values from physical priors. Both may allow for more complex priors and symmetries to be enforced instead of the simple vector calculus constraints we use here.

\newpage
\appendix 

\section{Notation and Conventions} \label{appendix: Notation}
\textbf{Bold} quantities are used to denote vectors. 

$\textbf{x}=\left(\begin{matrix}x\\y\\z\\\end{matrix}\right)$:  position. Non bold x,\ y\ and z are cartesian coordinates. 

$\textbf{v}=\ \left(\begin{matrix}v_x\\v_y\\v_z\\\end{matrix}\right)$: velocity vector

$\nabla\ =\left(\frac{d}{dx},\frac{d}{dy},\frac{d}{dz}\right)$: Gradient operator, also used in Divergence ($\nabla \bullet$) and Curl ($\nabla \times)$

\section{Lagrangian and Hamiltonian mechanics} \label{appendix: Lagrangian}
The Lagrangian of a particle in a magnetic field is:
\begin{equation}
    L=\frac{1}{2}mv^2-q\ V+q\textbf{A}\bullet\textbf{v} \label{eq: Magnetic Lagrangian}
\end{equation}
and the Hamiltonian is:
\begin{equation}
    H=\frac{1}{2}m \left(\textbf{p}-q\textbf{A}\right)^2+q\ V
\end{equation}

The Magnetic LNN's used in Experiment 1b are calculated using Lagrange's equations of motion from Equation \ref{eq: Magnetic Lagrangian}

Dissipative forces can be represented in Lagrangian mechanics a few ways. A time dependent Lagrangian can be used. For example, for a 1D dampened simple harmonic oscillator, $F=\ -\frac{dV}{dx}-\gamma v$, the following time dependent Lagrangian produces the correct force, $L=e^\frac{t\gamma}{m}\left(\frac{m}{2}v^2-V\right)$. 
Other methods of representing dissipative forces include Rayleigh Dissipation functions and doubling the degrees of freedom. However, learning these Lagrangians is difficult. Learned time dependent Lagrangians would only work within the time duration of the training. Doubling degrees of freedom increases complexity or may result in non-trivial Lagrangians that are hard to interpret being learned. 

\section{Constraints on types of usable Neural Networks}\label{appendix: Constraints}
We have created a series of neural network modules that can be selected to give the desired form of the equations of motion. However, care must be taken to ensure each module learns the correct aspect and each NN only learns its own part. Consider a model of the form: 

\begin{equation}
\textbf{F}=-\nabla V\left(\textbf{x}\right)+\textbf{f}\left(\textbf{v}\right)
\end{equation}
The learned neural networks can incorporate any constant vector $\textbf{h}$ that cancels leaving the overall force unchanged:
\begin{equation}
\textbf{F}^\prime=-\nabla\left[V^\prime\left(\textbf{x}\right)+\textbf{x}\bullet \textbf{h}\right]+\left[\textbf{f}^\prime\left(\textbf{v}\right)+\textbf{h}\right]=\textbf{F}
\end{equation}
The new learned velocity force $f^\prime\left(\textbf{v}\right)$ is shifted and new potential $V^\prime\left(\textbf{x}\right)$ are distorted by the addition of $x\bullet h$. However, the overall force remains unchanged, so the model’s predictions are identical. If we wanted the real potential or velocity, the returned quantities would be wrong. 

The most general equation of motion with separate drag and conservative velocity forces would be 
\begin{equation}
\textbf{F}= -\nabla V\left(\textbf{x}\right)+\widehat{\textbf{v}} \  \textbf{f}\left(\textbf{x}, \textbf{v}\right)+ \textbf{g}\left(\textbf{x}, \textbf{v}\right)\times\widehat{\textbf{v}} = \textbf{G}(\textbf{x},\textbf{v}),
\end{equation}
where any function of $\textbf{x}$ can be passed between the terms. This would mean the learned potential could be anything, the modules become uninterpretable, even if the overall model gives the correct predictions. This cannot be fixed without prior knowledge of the forces, so care must be taken when selecting modules.

\section{Experimental Setup} \label{appendix: Setup}
The models are trained as Neural ODEs on one body problems with time-independent forces. The dataset is generated by integrating the true equations of motion from randomly generated initial conditions in phase space. The simulations are run for time $t$, and loss is applied at $l$ reglularly spaced intervals. From the same initial conditions, the model is integrated to give predicted positions at each sample time. The loss is the mean squared error (MSE) between the predicted and true positions at each given time:
\begin{equation}
L_{pred}=\frac{1}{l}\sum_{n=1}^{l}\left|\textbf{x}_{pred} \left(\frac{nt}{l}\right)-\textbf{x}_{true} \left(\frac{nt}{l}\right)\right|^2.
\end{equation}
Adding any regularisation yields the overall loss $L=a_1L_{pred}+\sum_{i}{a_iL_i}$, where $a_i$ are learning rates and $L_i$ are regularisers, e.g. Eq. \ref{eq: div loss}. The weights are updated by backpropagation through the ODE solver. The models are evaluated in terms of the MSE over a longer simulation time starting from set initial conditions. The test conditions were chosen so the particle always remains within the train phase space.

Models were trained in batches of 8 at a time using Pythons multiprocessing on an 8 core AMD Ryzen 3700x
(ref), and all training times quoted are the time taken to train 8 models.

General parameters used for all tests:
\begin{itemize}
    \item Optimiser: Adam \cite{ADAM}
    \item Training steps: 16000
    \item Sequence length: 2
    \item Simulation training time: 0.2
    \item Learning rate decay on plateau with coefficient 0.8, patience 960
    \item Training phase space: $\textbf{x}=[-2, 2]^3, \textbf{v}=[-2, 2]^3$
    \item Simulation time for testing: 7
    \item Sequence length for test: 70
    \item NODE solver, dopri5 with rtol=0.003. Test are done using torchdiffeq Pytorch library \citep{Chen_torchdiffeq_2021}. All other solver parameters are default.
\end{itemize}

Each Neural Network module / model is built from fully connected multi-layer perceptions with Softplus activation \cite{softplus}. The shape and sizes of each module was chosen after testing several parameters to give the best performance. A summary of the NNs used to represent each module is given in Table \ref{tab:Model Training Summary}. 

The potential, magnetic and drag forces used were chosen to be reasonably complicated separate the performance of different types of models. They do not correspond directly to any well known physical forces. They were chosen by hand to have nice properties such as confining the particle and be difficult to learn. 

\begin{table*}
  \caption{Parameters of neural networks, showing the shape of each component as well as learning rate used and total
number of trainable parameters. Numbers in dashes represent number of hidden neurons, with the first and last number being input and output size,
respectively. In Magnetic and Div. Magnetic, the magnetic fields was directly learned directly, while in LNN and Vector
Magnetic, the vector potential was learned. The learning rate for regularisation terms is shown for the Divergence Magnetic NN.}
  \label{tab:Model Training Summary}
  \centering
  \begin{tabular}{llllll}
    \toprule
    \cmidrule(r){1-2}
    Model          & Potential        & Resistance   & Magnetic field   & Learning rate & Parameters \\
    \midrule
    Magnetic NN    & 3-25-25-25-1     & 3-25-25-25-1 & 3-25-25-25-3     & 15e-3         & 4325\\
    Div. Mag.NN    & 3-25-25-25-1     & 3-25-25-25-1 & 3-25-25-25-3     & 15e-3 + 2e-7  & 4325\\
    Vector NN      & 3-25-25-25-1     & 3-25-25-25-1 & 3-25-25-25-3     & 10e-3         & 4325\\
    Basic    NN    & 3-40-40-40-1     & \multicolumn{2}{c}{$6-40-40-40-3$}& 5e-3         & 7160\\
    Periodic NN    & 3-25-25-25-1     & 3-25-25-25-1 & 3-25-25-25-3     & 8e-3         & 4325\\
    Magnetic LNN   & 3-25-25-25-1     &              & 3-25-25-25-3     & 10e-3         & 2900\\
    LNN            &  \multicolumn{3}{c}{$6-35-35-35-1$}                & 4e-3         & 2800\\
    SONODE         &  \multicolumn{3}{c}{$6-120-120-120-3$}             & 1e-3         & 44760\\

    \bottomrule
  \end{tabular}
\end{table*}
\subsection{Standard Test Setup}
This standard test setup used in experiment 1a consists of: 
Standard potential:
\begin{equation}
    V(\textbf{x}) = (4 sin(z) * y + x) r e^{-1.5 r ^2} +e^{-3 r^2} + \frac{1}{\left(1 + e^{-4 r + 8} \right)}
\end{equation}    
\begin{equation}
        r= \sqrt{x^2 + y^2 + z^2}
\end{equation}

Standard magnetic field:
\begin{equation}
    \textbf{B} = \begin{pmatrix} x\ z \\ x\ cos(z) \\ -\frac{z^2}{2} + sin(y) \end{pmatrix}
\end{equation}

Standard Drag:

\begin{equation}
    \textbf{F}_d = -\textbf{v}D(\textbf{v})
\end{equation}
\begin{equation}
    D(\textbf{v}) = \frac{1}{2} v_r^2 e^{-v_r / 3}\ sin(\theta) cos(\theta) sin(\phi) - 1
\end{equation}
\begin{equation}
    v_r = \sqrt{v_x^2 + v_y^2 + v_z^2},
\end{equation}
\begin{equation}
\phi=arctan(\frac{v_y}{v_x}), \theta=arcos(\frac{v_z}{v_r + 0.01})
\end{equation}    

In Experiment 1b, the drag force was removed. Everything else remained the same. 

\subsection{Magnetic Testing setup}
The magnetic testing setup in Experiment 2 uses a more complex magnetic field and simpler potential. The drag is the Standard drag. Potential is 
\begin{equation}
    V = r^2.
\end{equation}
Magnetic field is:
\begin{equation}
    \textbf{B} = \begin{pmatrix} x\ z + cos(\frac{y}{2})\\ x\ cos(z) + 0.5 x\ cos(y)\\ -\frac{z^2}{2} + 0.5\ x\ z\ sin(y) \end{pmatrix}
\end{equation}

\subsection{Periodic Potential Setup}
In Experiment 3, testing periodic potentials, the setup is the same as the \textit{Standard} Test Setup, except the potential is replaced with a periodic potential
\begin{equation}
\begin{aligned}
    V(\textbf{x}) = cos(x \pi) -  sin\left((x+y) \pi\right) + cos\left((x+z) \pi\right) + 
    \\ 0.9 cos(y \pi) + 0.7 cos\left((y + z) \pi)\right) - sin(z \pi) 
    \\- cos\left((x-y+z)\pi\right) 
    - sin\left((-x+z)\pi)\right)
\end{aligned}
\end{equation}

\subsection{Combining Models}
In Experiment 4, $\mathcal{M}_A$ and $\mathcal{M}_B$ are trained using the \textit{Magnetic}
Testing setup and \textit{Standard} Test Setup respectively. $\mathcal{M}_C$ was built by taking the learned NN functions and combining them into a third model.

\FloatBarrier

\bibliography{main}
\bibliographystyle{aaai}

\newpage

\end{document}